\documentclass[preprint,12pt]{elsarticle}

\usepackage{amssymb}
\usepackage{multirow}
\usepackage{booktabs}
\usepackage{mathrsfs}
\usepackage{longtable}
\usepackage{lscape}
\usepackage{tabularx}
\usepackage{epsfig}
\usepackage{colortbl}
\usepackage{slashbox}
\usepackage{dcolumn,longtable,hhline,colortbl}
\usepackage[table]{xcolor}
\usepackage[rotateright]{rotating}
\usepackage{multirow}
\usepackage{morefloats}
\usepackage{amsthm} 
\usepackage{longtable,lscape}

\newdefinition{definition}{Definition}

\journal{arXiv.org}

\begin{document}

\begin{frontmatter}



\title{Transformation of basic probability assignments to probabilities based on a new entropy measure}


\author[swu,CQS]{Xinyang Deng}
\author[swu,vu]{Yong Deng\corref{cor}}
\ead{prof.deng@hotmail.com}

\address[swu]{School of Computer and Information Science, Southwest University, Chongqing, 400715, China}
\address[CQS]{Center for Quantitative Sciences, Vanderbilt University School of Medicine, Nashville, TN, 37232, USA}
\address[vu]{Department of Civil \& Environmental Engineering, School of Engineering, Vanderbilt University, Nashville, TN, 37235, USA}

\cortext[cor]{Corresponding author: Yong Deng, School of Computer and Information Science, Southwest University, Chongqing, 400715, China.}

\begin{abstract}
Dempster-Shafer evidence theory is an efficient mathematical tool to deal with uncertain information. In that theory, basic probability assignment (BPA) is the basic element for the expression and inference of uncertainty. Decision-making based on BPA is still an open issue in Dempster-Shafer evidence theory. In this paper, a novel approach of transforming basic probability assignments to probabilities is proposed based on Deng entropy which is a new measure for the uncertainty of BPA. The principle of the proposed method is to minimize the difference of uncertainties involving in the given BPA and obtained probability distribution. Numerical examples are given to show the proposed approach.
\end{abstract}

\end{frontmatter}

\section{Introduction}\label{Introduction}
Dempster-Shafer evidence theory is widely used in many disciplines since it allows to deal with uncertain information. Several familiar branches of its applications includes statistical learning \cite{cuzzolin2008geometric,huang2014new,yang2013novel}, classification and clustering \cite{denoeux1995k,denoeux2000neural,masson2008ecm,liu2013evidential}, decision making \cite{liu2014PM,ahn2014use,yao2014induced}, knowledge reasoning \cite{kang2012evidential,denoeux2013maximum}, risk assessment and evaluation \cite{zhang2013ifsjsp,Deng2014DAHPSupplier,yager2014characterizing}, and so forth \cite{chen2013fuzzy,deng2014environmental,zhang2014wei,yager2014owa,deng2015parameter}. In Dempster-Shafer evidence theory, several key research directions continuingly appeal to researcher's attention, for example, the combination of multiple evidences \cite{murphy2000combining,lefevre2013preserve,deng2014improved}, conflict management \cite{liu2006analyzing,schubert2011conflict}, generation of basic probability assignment (BPA) \cite{xu2014nonAppInt,zhang2014new,xu2013newKBS}, and so on \cite{karahan2013persistence,certa2013multistep,zhang2014response}. Among these points, decision-making based BPA is a crucial issue to be solved, and it has attracted much attention.

A lot of works have been done to construct a reasonable model for the decision making based on the BPA \cite{smets1994transferable,smets2005decision,merigo2011decision,nusrat2013descriptive}. One widely used model is the transferable belief model (TBM) \cite{smets1994transferable}, pignistic probabilities are used for decision making in this model. In the TBM, a pignistic probability transformation (PPT) approach has been proposed to bring out probabilities from BPAs. Another well-known probability transformation method is proposed by Barry R. Cobb \cite{cobb2006plausibility}, which is based on the plausibility function. The main idea of the plausibility transformation method is to assign the uncertain according to the plausibility function with normalization. In \cite{cuzzolin2012relative}, the semantics and properties of the relative belief transform have been discussed. One method was mentioned namely proportional probability transformation \cite{daniel2006transformations}. Within the proportional probability transformation, a belief mass assigned to nonsingleton focal element $X$ is distributed among $X$'s elements with respect influenced by the proportion of BPAs assigned to singletons. The proportional probability transformation is influenced by the proportion of BPAs assigned to singletons.

In this paper, a novel probability transformation approach is proposed based on a new entropy measure of BPAs, Deng entropy \cite{DengEntropyArXiv}. Within the proposed approach, given a BPA, it is expected to find a probability distribution whose Shannon entropy is as close as possible to the entropy of given BPA. The rest of this paper is organized as follows. Section \ref{Preliminaries} introduces some basic background knowledge. In section \ref{ProposedMethod} the proposed probability transformation approach is presented. Section \ref{NumExample} uses some examples to illustrate the effectiveness of the proposed approach. Conclusion is given in Section \ref{Conclusion}.

\section{Preliminaries}\label{Preliminaries}

\subsection{Dempster-Shafer evidence theory}

Dempster-Shafer evidence theory \cite{dempster1967upper,shafer1976mathematical}, also called Dempster-Shafer theory or evidence theory, is used to deal with uncertain information. As an effective theory of uncertainty reasoning, Dempster-Shafer theory has an advantage of directly expressing various uncertainties. This theory needs weaker conditions than bayesian theory of probability, so it is often regarded as an extension of the bayesian theory. For completeness of the explanation, a few basic concepts are introduced as follows.

\begin{definition}\label{b1}
Let $\Omega$ be a set of mutually exclusive and collectively
exhaustive, indicted by
\begin{equation}\label{q_1}
\Omega  = \{ E_1 ,E_2 , \cdots ,E_i , \cdots ,E_N \}
\end{equation}
The set $\Omega$ is called frame of discernment. The power set of
$\Omega$ is indicated by $2^\Omega$, where
\begin{equation}\label{q_2}
2^\Omega   = \{ \emptyset ,\{ E_1 \} , \cdots ,\{ E_N \} ,\{ E_1
,E_2 \} , \cdots ,\{ E_1 ,E_2 , \cdots ,E_i \} , \cdots ,\Omega \}
\end{equation}
If $A \in 2^\Omega$, $A$ is called a proposition.
\end{definition}

\begin{definition}\label{b2}
For a frame of discernment $\Omega$,  a mass function is a mapping
$m$ from  $2^\Omega$ to $[0,1]$, formally defined by:
\begin{equation}\label{q_3}
m: \quad 2^\Omega \to [0,1]
\end{equation}
which satisfies the following condition:
\begin{eqnarray}\label{q_4}
m(\emptyset ) = 0 \quad \rm{and} \quad \sum\limits_{A \in 2^\Omega }
{m(A) = 1}
\end{eqnarray}
\end{definition}

In Dempster-Shafer theory, a mass function is also called a basic
probability assignment (BPA). If $m(A) > 0$, $A$ is called a focal
element, the union of all focal elements is called the core of the
mass function.

\begin{definition}\label{b3}
For a proposition $A \subseteq \Omega$, the belief function
$Bel:\;2^\Omega   \to [0,1]$ is defined as
\begin{equation}\label{Belfunction}
Bel(A) = \sum\limits_{B \subseteq A} {m(B)}
\end{equation}
The plausibility function $Pl:\;2^\Omega   \to [0,1]$ is defined
as
\begin{equation}\label{Plfunction}
Pl(A) = 1 - Bel(\bar A) = \sum\limits_{B \cap A \ne \emptyset }
{m(B)}
\end{equation}
where $\bar A = \Omega - A$.
\end{definition}

Obviously, $Bel(A) \le Pl(A)$, these functions $Bel$ and $Pl$ are
the lower limit function and upper limit function of proposition
$A$, respectively.

\section{Proposed probability transformation approach based on Deng entropy}\label{ProposedMethod}
In this section, a new measure for the uncertainty of BPA, Deng entropy is introduced first, then a new approach of transforming BPA to probability distribution is proposed based on the concept of Deng entropy.

\subsection{Deng entropy}
Deng entropy \cite{DengEntropyArXiv} is a generalized Shannon entropy to measure uncertainty involving in a BPA. Mathematically, Deng entropy can be presented as follows
\begin{equation}
E_d  =  - \sum\limits_i {m(F_i )\log \frac{{m(F_i )}}{{2^{|F_i|}  - 1}}}
\end{equation}
where, $F_i$ is a proposition in mass function $m$, and $|F_i|$ is the cardinality of $F_i$. As shown in the above definition, Deng entropy, formally,  is similar with the classical Shannon entropy, but the belief for each proposition $F_i$ is divided by a term $({2^{|F_i|}  - 1})$ which represents the potential number of states in $F_i$ (of course, the empty set is not included).

Specially, Deng entropy can definitely degenerate to the Shannon entropy if the belief is only assigned to single elements. Namely,
\begin{equation}
  E_d  =  - \sum\limits_i {m(\theta _i )\log \frac{{m(\theta _i )}}{{2^{|\theta _i|}  - 1}}}  =  - \sum\limits_i {m(\theta _i )\log m(\theta _i )}
\end{equation}

\subsection{Proposed probability transformation approach}
In our view, a primary principle in the transformation process is to minimize the difference of uncertainties involving in the given BPA and obtained probability distribution. In order to implement such optimization transformation, it must be able to calculate the uncertainty of BPA. Exactly, Deng entropy provides a method to measure the uncertainty of BPA as well as probability distribution. Therefore, a novel probability transformation approach based on Deng entropy can be proposed as follows.

Assume the frame of discernment is $\Omega = \{\omega_1, \omega_2, \cdots, \omega_n\}$, given a BPA $m$, a probability distribution $P = \left( p(\omega_1), p(\omega_2), \cdots, p(\omega_n)\right)$ associated with $m$ is calculated by solving the following optimization problem:

\begin{equation}
\begin{array}{l}
 \min \quad \left| {E_d (m) - E_d (P)} \right| \\
 s.t.\quad \left\{ \begin{array}{l}
 \sum\limits_i^n {p(\omega_i) }  = 1; \\
 Bel(\omega _i ) \le p(\omega_i)  \le Pl(\omega _i ),\quad i = 1, \cdots ,n. \\
 \end{array} \right. \\
 \end{array}
\end{equation}
where $E_d (m)$ and $E_d (P)$ are the entropies of BPA $m$ and probability distribution $P$, respectively.

\section{Numerical examples}\label{NumExample}
In this section, some illustrative examples are given to show the proposed probability transformation approach.

\textbf{Example 1.}
Given a frame of discernment $\Omega = \{\omega_1, \omega_2, \omega_3, \omega_4\}$, there is a BPA $m(\omega_1,\omega_2,\omega_3,\omega_4) = 1$. According to Eqs. (\ref{Belfunction}) and (\ref{Plfunction}),

$Bel(\omega _1 ) = Bel(\omega _2 ) = Bel(\omega _3 ) = Bel(\omega _4 ) = 0$,

$Pl(\omega _1 ) = Pl(\omega _2 ) = Pl(\omega _3 ) = Pl(\omega _4 ) = 1$.

By using the proposed probability transformation approach, a probability distribution is obtained by
\[
\begin{array}{l}
 \min \quad \left| {E_d (m) - E_d (P)} \right| \\
 s.t.\quad \left\{ \begin{array}{l}
 p(\omega _1 ) + p(\omega _2 ) + p(\omega _3 ) + p(\omega _4 ) = 1 \\
 0 \le p(\omega _i ) \le 1,\quad i = 1,2,3,4. \\
 \end{array} \right. \\
 \end{array}
\]

we can obtain that 

$P: p(\omega _1 ) = 0.25, p(\omega _2 ) = 0.25, p(\omega _3 ) = 0.25, p(\omega _4 ) = 0.25.$

The result shows that the transformed probability distribution has the maximum uncertainty (Shannon entropy) when the given BPA is totally unknown (i.e., $m(\Omega)=1$).

%
%

\textbf{Example 2.}
Given a frame of discernment $\Omega = \{\omega_1, \omega_2, \omega_3\}$, there is a BPA: $m(\omega_1)=0.4, m(\omega_2)=0.05, m(\omega_3)=0.1, m(\omega_1, \omega_2)=0.1, m(\omega_1, \omega_3)=0.2, m(\omega_1, \omega_2, \omega_3)=0.15.$

Due to 
$Bel(\omega _1 ) = 0.4,Bel(\omega _2 ) = 0.05,Bel(\omega _3 ) = 0.1$; $Pl(\omega _1 ) = 0.85,Pl(\omega _2 ) = 0.3,Pl(\omega _3 ) = 0.45$, the associated probability distribution can be calculated by
\[
\begin{array}{l}
 \min \quad \left| {E_d (m) - E_d (P)} \right| \\
 s.t.\quad \left\{ \begin{array}{l}
 p(\omega _1 ) + p(\omega _2 ) + p(\omega _3 ) = 1 \\
 0.4 \le p(\omega _1 ) \le 0.85 \\
 0.05 \le p(\omega _2 ) \le 0.3 \\
 0.1 \le p(\omega _3 ) \le 0.45 \\
 \end{array} \right. \\
 \end{array}
\]
So, we can get $P: p(\omega _1 ) = 0.4, p(\omega _2 ) = 0.3, p(\omega _3 ) = 0.3.$

\section{Conclusion}\label{Conclusion}
In this paper, the transformation of BPA to probability distribution has been studied. Based on an idea that minimizing the difference of uncertainties involving in the given BPA and obtained probability distribution, a novel probability transformation approach has been proposed. Finally, several illustrative examples have been given to show the proposed method.

\section*{Acknowledgements}
The work is partially supported by China Scholarship Council, National Natural Science Foundation of China (Grant No. 61174022), Specialized Research Fund for the Doctoral Program of Higher Education (Grant No. 20131102130002), R\&D Program of China (2012BAH07B01), National High Technology Research and Development Program of China (863 Program) (Grant No. 2013AA013801), the open funding project of State Key Laboratory of Virtual Reality Technology and Systems, Beihang University (Grant No.BUAA-VR-14KF-02), Fundamental Research Funds for the Central Universities (Grant No. XDJK2014D034).

%



\bibliographystyle{elsarticle-num}
\bibliography{reference}







\end{document}